\documentclass{article}  

\usepackage{hyperref} 
\usepackage{graphicx}
\usepackage{amssymb,amsfonts,amsmath,dsfont,amsthm}

\usepackage[utf8]{inputenc} % allow utf-8 input
\usepackage[T1]{fontenc}    % use 8-bit T1 fonts
\usepackage{url}            % simple URL typesetting
\usepackage{booktabs}       % professional-quality tables
\usepackage{nicefrac}       % compact symbols for 1/2, etc.
\usepackage{microtype}      % microtypography    
\usepackage{graphicx}

\usepackage[svgnames]{xcolor}
\usepackage{xcolor,colortbl}

\providecommand{\scal}[2]{\left\langle{#1},{#2}\right\rangle}    
\newcommand{\C}{\mathcal{C}}

\renewcommand{\O}{\mathcal{O}}

\DeclareMathOperator{\R}{\mathbb{R}}

\providecommand{\scal}[2]{\left\langle{#1},{#2}\right\rangle}

\newcommand{\be}{\begin{equation}}
\newcommand{\ee}{\end{equation}}

\newcommand{\bt}{\begin{theorem}}
\newcommand{\et}{\end{theorem}}
\newcommand{\bd}{\begin{definition}}
\newcommand{\ed}{\end{definition}}
\newcommand{\br}{\begin{remark}}
\newcommand{\er}{\end{remark}}
\newcommand{\x}{\mathbf{x}}
\newcommand{\y}{\mathbf{y}}

\renewcommand{\k}{\mathbf{k}}
\newcommand{\XX}{\mathbb{X}}
\newtheorem{theorem}{Theorem}
\newtheorem{definition}{Definition}

\newtheorem{remark}{Remark}

\usepackage[margin= 0.8in,top=12mm]{geometry}        
\newcommand*{\titleAT}{\begingroup
  \newlength{\drop}
  \drop=0.05\textheight

  \includegraphics[scale=1.5]{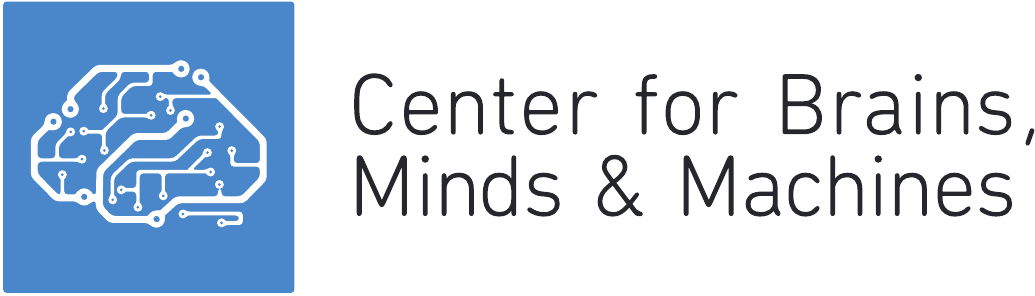}

  \textcolor{CornflowerBlue}{\rule{\textwidth}{3 pt}}\par
  \vspace{2pt}\vspace{-\baselineskip}
  \rule{\textwidth}{0.4pt}\par

  \vspace{\drop}
  \textbf{\large{CBMM Memo No. \memonumber}}   \hfill    \textbf{\large{\memodate}} 

  \vspace{\drop}
  \begin{center}
    \textbf{\huge{\memotitle}}\\
    \vspace{0.4\drop}
    \textbf{\Large{by}}\\
    \vspace{0.4\drop}
    \large{\memoauthors}
  \end{center}
  \vspace{\drop}
  \textbf{\large{\noindent Abstract}:} {\memoabstract}

  \textcolor{CornflowerBlue}{\rule{\textwidth}{3 pt}}\par
  \vspace{2pt}\vspace{-\baselineskip}
  \rule{\textwidth}{0.4pt}\par 
  
  \begin{minipage}{.15\linewidth}
    \includegraphics[scale=0.1]{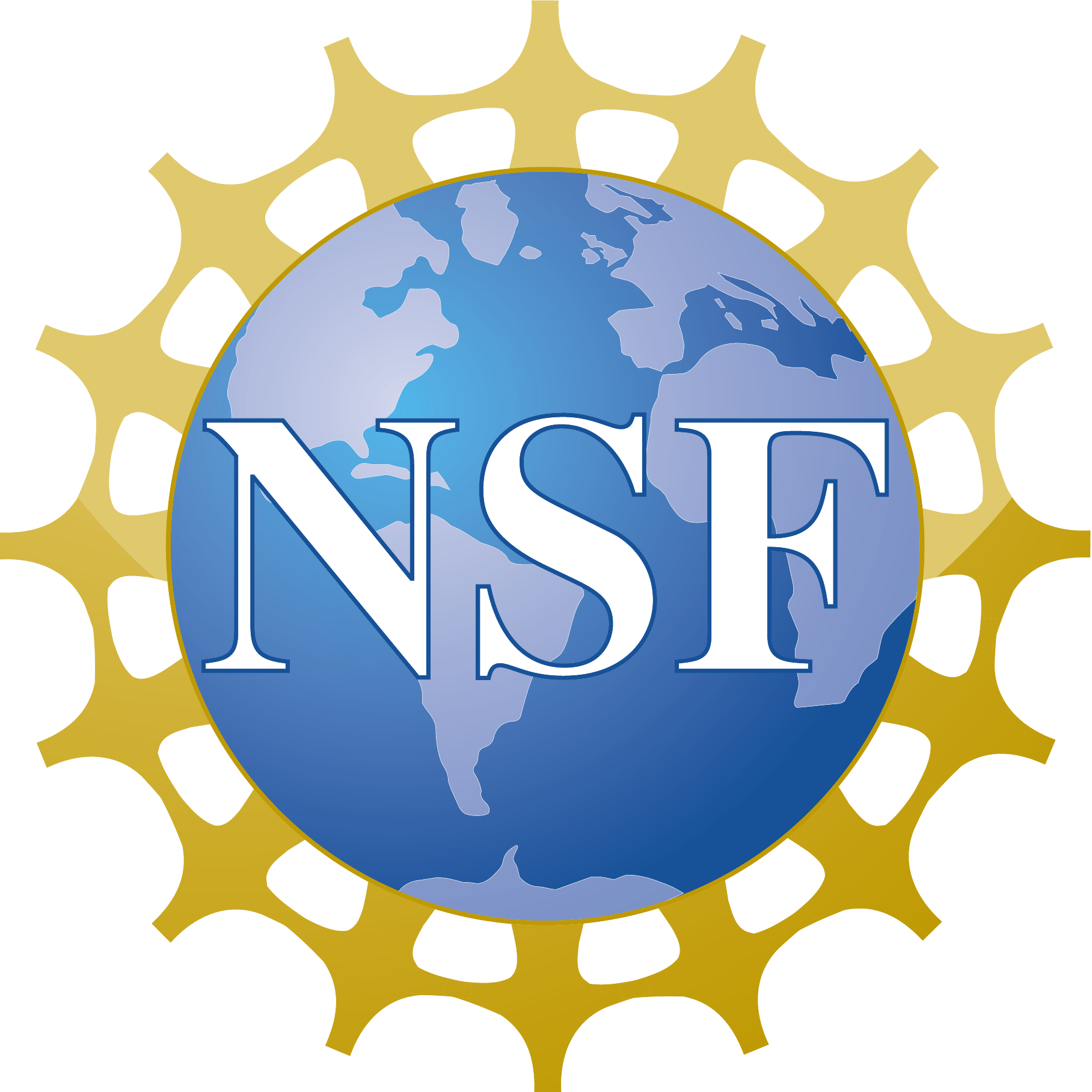}
  \end{minipage}
  \begin{minipage}{.84\linewidth}
    \textbf{\large{This work was supported by the Center for Brains, Minds and Machines (CBMM), funded by NSF STC award  CCF - 1231216.}}
  \end{minipage}

  \endgroup} 

\begin{document}

%% %%% 
%% \maketitle
%% \begin{abstract}
%%   While the universal approximation property holds both for
%%   hierarchical and shallow networks, we prove that deep (hierarchical)
%%   networks can approximate the  class of {\it compositional functions}
%%   with the same accuracy as shallow networks but with exponentially
%%   lower number of training parameters as well as VC-dimension. This
%%   theorem settles an old conjecture by Bengio on the
%%   role of depth in networks. We then define a general class of scalable,
%%   shift-invariant algorithms to show a simple and natural set of
%%   requirements that  justify deep convolutional networks.
%% \end{abstract}
%% %%

%%%%%%%%%%%%%%%%% MEMO START %%%%%%%%%%%%%%%%%%%%%%
\def\memonumber{045}
\def\memodate{\today}     
\def\memotitle{Learning Functions: When Is Deep Better Than Shallow}
\def\memoauthors{\textbf{Hrushikesh Mhaskar} \\  
  Department of Mathematics, California Institute of Technology, Pasadena, CA 91125;\\
  Institute of Mathematical Sciences, Claremont Graduate University, Claremont, CA 91711,\\
  \textbf{Qianli Liao and Tomaso Poggio} \\  
  Center for Brains, Minds, and Machines, McGovern Institute for Brain Research \\
  Massachusetts Institute of Technology, Cambridge, MA, 02139 \\  
}
\def\memoabstract{\footnote{This is an update of a previous version released on March 3rd 2016.} 
  While the universal approximation property holds both for
  hierarchical and shallow networks, we prove that deep (hierarchical)
  networks can approximate the  class of {\it compositional functions} 
  with the same accuracy as shallow networks but with exponentially
  lower number of training parameters as well as VC-dimension. This
  theorem settles an old conjecture by Bengio on the
  role of depth in networks. We then define a general class of scalable,
  shift-invariant algorithms to show a simple and natural set of
  requirements that  justify deep convolutional networks.
} 

\titleAT

\newpage

%%%%%%%%%%%%%%%%%%%%%%%%%%%%%%%%%%%%%%%%%%%%%%% MEMO END %%%%%%%%%%%%%

\section{Introduction}

There are two main theory questions about Deep Neural Networks.  The
first question is about the power of the architecture -- which classes
of functions can it approximate well?  The second question is about
learning the unknown coefficients from the data: why is SGD so
unreasonably efficient, at least in appearance?  Are good minima
easier to find in deep rather than in shallow networks?  In this paper
we describe a set of approximation theory results that include answers
to why and when deep networks are better than shallow by using the
idealized model of a deep network as a binary tree. In a separate
paper, we show that the binary tree model with its associated results
can  indeed be extended formally to the very deep convolutional
networks of the ResNet type which have only a few stages of pooling
and subsampling.

This paper compares shallow (one-hidden layer) networks
with deep networks (see for example Figure
\ref{example_function_8variables}).  Both types of networks use the
same small set of operations -- dot products, linear combinations, a
fixed nonlinear function of one variable, possibly convolution and
pooling. The logic of the paper is as follows.

\begin{itemize}

\item Both shallow (a) and deep (b) networks  are {\it universal},
  that is they can approximate arbitrarily well any continuous
  function of $d$ variables on a compact domain.

\item We show that the approximation of functions with a {\it
    compositional structure} -- such as $f(x_1, \cdots, x_{d}) = h_{1}
  (h_{2} \cdots (h_j (h_{i1} (x_1, x_2), h_{i2}(x_3, x_4)), \cdots)) $
  -- can be achieved with the same degree of accuracy by deep and
  shallow networks but that the number of parameters, the VC-dimension and the fat-shattering
  dimension are much smaller for the deep networks than for the
  shallow network with equivalent approximation accuracy. It is
  intuitive that a hierarchical network matching the structure of a
  compositional function should be ``better'' at approximating it than
  a generic shallow network but universality of shallow networks makes
  the statement less than obvious.  Our result makes clear that the
  intuition is indeed correct and provides quantitative bounds.

\item Why are compositional functions important? We argue that the
  basic properties of scalability and shift invariance in many natural
  signals such as images and text require compositional algorithms
  that can be well approximated by Deep Convolutional Networks. Of
  course, there are many situations that do not require shift
  invariant, scalable algorithms. For the many functions that are not
  compositional we do not expect any advantage of deep convolutional
  networks.
\end{itemize}

\section{Previous work}

The success of Deep Learning in the present landscape of machine
learning poses again an old theory question:
why are multi-layer networks better than one-hidden-layer networks?
Under which conditions? The question is relevant in several related
fields from machine learning to function approximation and has
appeared many times before. 

Most Deep Learning references these days start with Hinton's
backpropagation and with Lecun's convolutional networks (see for a
nice review \cite{LeCunBengioHinton2015}). Of course, multilayer
convolutional networks have been around at least as far back as the
optical processing era of the 70s. The Neocognitron
(\cite{fukushima:1980}) was a convolutional neural network that was
trained to recognize characters. The HMAX model of visual cortex
(\cite{riesenhuber:1999}) was described as a series of AND and OR
layers to represent hierarchies of disjunctions of conjunctions.
There are several recent papers addressing the question of why
hierarchies.  Sum-Product networks, which are equivalent to polynomial
networks (see \cite{Moore1988,DBLP:journals/corr/abs-1304-7045}), are
a simple case of a hierarchy that can be analyzed (\cite{DBLP:conf/nips/DelalleauB11}).  \cite{MontufarBengio2014}
provided an estimation of the number of linear regions that a network
with ReLU nonlinearities can synthesize in principle but leaves open
the question of whether they can be used for learning. Examples of
functions that cannot be represented efficiently by shallow networks
have been given very recently by \cite{Telgarsky2015}.  Most relevant
to this paper is the work on hierarchical quadratic networks
(\cite{DBLP:journals/corr/abs-1304-7045}), together with function
approximation results (\cite{Pinkus1999,mhaskar1993neural}).

\section{Compositional functions}

We assume that the shallow networks do not have any structural
information on the function to be learned (here its compositional
structure), because they cannot represent it directly. Deep networks
with standard architectures on the other hand {\it do represent}
compositionality and can be adapted to the details of such prior
information. Thus, it is natural to conjecture that hierarchical compositions of functions such as
\begin{eqnarray}
  \lefteqn{f(x_1,
  \cdots, x_8) = h_3(h_{21}(h_{11} (x_1, x_2),  h_{12}(x_3, x_4)),}\nonumber\\
 && \qquad h_{22}(h_{13}(x_5, x_6),  h_{14}(x_7, x_8)))
 \label{l-variables}
\end{eqnarray}
\noindent are approximated more efficiently by deep than by shallow networks. 

In addition, both shallow and deep representations may or may not
reflect invariance to group transformations of the inputs of the
function  ( \cite{Soatto2011,anselmi2015theoretical}). Invariance is expected to
decrease the complexity of the network, for instance its VC-dimension.
Since we are interested in the comparison of shallow vs deep
architectures, here we consider the generic case of networks (and
functions) for which invariance is not assumed.

We approximate functions of $d$ variables of the form of Equation
(\ref{l-variables})  with networks in which the activation
nonlinearity is a smoothened version of the so called ReLU, originally called {\it ramp} by
Breiman and given by $\sigma (x) = x_+ = max(0, x)$ .
The architecture of the deep networks reflects Equation
(\ref{l-variables}) with each node $h_i$ being a ridge function.

\begin{figure}\centering
\includegraphics[width=0.6\textwidth]{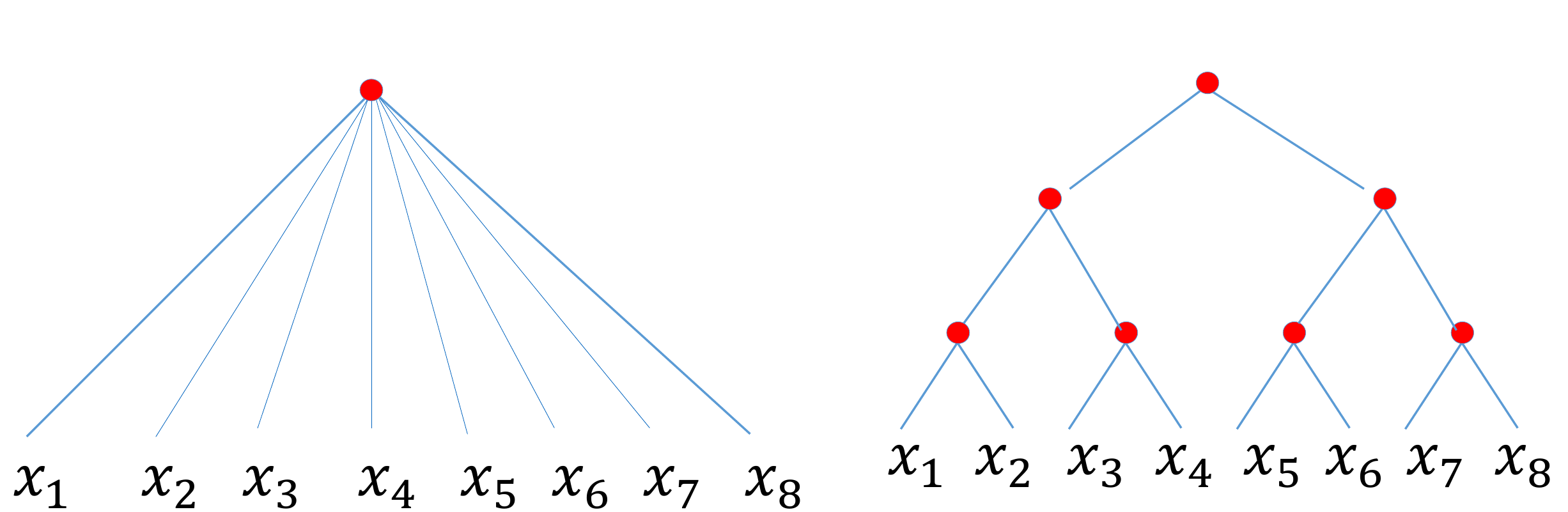}
\caption{ a) A shallow universal network in 8 variables and $N$ units
  which can approximate a generic function $f(x_1, \cdots, x_8)$. b) A
  binary tree hierarchical network in 8 variables, which approximates
  well functions of the form $f(x_1, \cdots, x_8) = h_3(h_{21}(h_{11}
  (x_1, x_2), h_{12}(x_3, x_4)), \allowbreak h_{22}(h_{13}(x_5, x_6),
  h_{14}(x_7, x_8))) $.  Each of the nodes in b) consists of $n$ ReLU
  units and computes the ridge function   (\cite{Pinkus1999})
  $\sum_{i=1}^n a_i(\scal{\mathbf{v}_i}{\mathbf{x}}+t_i)_+$, with
  $\mathbf{v}_i, \mathbf{x} \in \R^2$, $a_i, t_i\in\R$. Each term,
  that is each unit in the node, 
  corresponds to a ``channel''. Similar to the shallow network a
  hierarchical network as in b) can approximate any continuous function;
  the text proves how it approximates a compositional functions better
  than a shallow network.  No invariance is assumed here.}
\label{example_function_8variables}
\end{figure}

It is {\it important to emphasize} here that the properties of state-of-art Deep
Learning Neural Networks (DLNNs) of the ResNet type
(\cite{EXTREMELYDEEP_MS_2015}), with their small kernel size and many
layers, are  well represented by our results on binary tree architectures,
as we show formally elsewhere. Visual cortex has a similar
compositional architecture with receptive fields becoming larger and
larger in higher and higher visual areas, with each area corresponding
to a recurrent layer in a deep neural network (\cite{ LiaoPoggio2016}).

\section{Main results}\label{approxsect}

In this section, we describe the approximation properties of the shallow and deep
networks in two cases: deep networks with ReLU nonlinearities and deep
Gaussian networks. The general paradigm is as follows. We are interested in
determining how complex the network ought to be to
\textbf{theoretically guarantee} that the network would approximate an
unknown target function $f$ up to a given accuracy $\epsilon>0$.  To
measure the accuracy, we need a norm $\|\cdot\|$ on some normed linear
space $\mathbb{X}$.  The complexity of the network is indicated by a
subscript to the general class of networks with this complexity; let
$V_n$ the be set of all networks of a given kind with the complexity
given by $n$ (e.g., all shallow networks with $n$ units in the
hidden layer). It is assumed that the class of networks with a higher
complexity include those with a lower complexity; i.e., $V_n\subseteq
V_{n+1}$. The \textit{degree of approximation} is defined by
\begin{equation}
\label{degapproxdef}
\mathsf{dist}(f, V_n)=\inf_{P\in V_n}\|f-P\|.
\end{equation}
For example, if $\mathsf{dist}(f, V_n)=\O(n^{-\gamma})$ for some
$\gamma>0$, then one needs a network with complexity
$\O(\epsilon^{-\gamma})$ to guarantee an approximation up to accuracy
$\epsilon$. It is therefore customary and easier to give estimates on
$d(f, V_n)$ in terms of $n$ rather than in terms of  $\epsilon$, inverse to $d(f,V_n)$. Since $f$ is unknown, in order
to obtain theoretically proved upper bounds, we need to make some
assumptions on the class of functions from which the unknown target
function is chosen. This a priori information is codified by the
statement that $f\in W$ for some subspace $W\subseteq
\mathbb{X}$. This subspace is usually referred to as the smoothness
class.  In general, a deep network architecture (in this paper, we
restrict ourselves to the binary tree structure as in
(\ref{example_function_8variables})) has an advantage over the shallow
networks when the target function itself has the same hierarchical,
compositional structure, e.g., (\ref{example_function_8variables}).

\subsection{Deep and shallow neural networks}\label{degapproxsect}

We start with a review of previous results (\cite{optneur}).  Let
$I^d=[-1,1]^d$, $\XX=C(I^d)$ be the space of all continuous functions
on $I^d$, with $\|f\|=\max_{\x\in I^d}|f(\x)|$. Let $\mathcal{S}_n$
denote the class of all shallow networks with $n$ units of the form
$$
\x\mapsto\sum_{k=1}^n a_k\sigma(\mathbf{w}_k\cdot\x+b_k),
$$
where $\mathbf{w}_k\in\R^d$, $b_k, a_k\in\R$. The number of trainable
parameters here is $(d+2)n\sim n$. Let $r\ge 1$ be an integer, and
$W_{r,d}^{\mbox{NN}}$ be the set of all functions with continuous partial
derivatives of orders up to $r$ such  that $\|f\|+\sum_{1\le |\k|_1\le r} \|D^\k f\| \le 1$, where $D^\k$ denotes the partial derivative indicated by the
multi--integer $\k\ge 1$, and $|\k|_1$ is the sum of the components of
$\k$.

For the hierarchical binary tree network, the analogous spaces are defined by considering the compact set $W_{H,r,2}^{\mbox{NN}}$ to be the
class of all functions $f$ which have the same structure (e.g.,
(\ref{l-variables})), where each of the constituent functions $h$ is
in $W_{r,2}^{\mbox{NN}}$ (applied with only $2$ variables).  We define the corresponding
class of deep networks $\mathcal{D}_n$ to be set of all functions with
the same structure, where each of the constituent functions  is in
$\mathcal{S}_n$. We note that in the case when $d$ is an integer power of $2$, the number of parameters involved in an element of
$\mathcal{D}_{n}$ -- that is, weights and biases, in a node of the binary tree  is $(d-1)(d+2)n$.

The following theorem estimates the degree of approximation for shallow and deep networks. We remark that the assumptions on $\sigma$ in the theorem below are not satisfied by the
ReLU function $x\mapsto x_+$, but they are satisfied by smoothing the function in an
arbitrarily small interval around the origin. 

\begin{theorem}
\label{optneurtheo}
Let $\sigma :\R\to \R$ be infinitely differentiable, and not a polynomial on any subinterval of $\R$. \\
{\rm (a)} For $f\in W_{r,d}^{\mbox{NN}}$
\be\label{optneurest}
\mathsf{dist}(f,\mathcal{S}_n)= \O(n^{-r/d}).
\ee 
{\rm (b)} For $f\in W_{H,r,d}^{\mbox{NN}}$
\begin{equation}
\mathsf{dist}(f,\mathcal{D}_n) =\mathcal{O}(n^{-r/2}).
\label{deepnetapprox}
\end{equation}
\end{theorem}

\noindent\textit{Proof.}
Theorem~\ref{optneurtheo}(a) was proved by
  \cite{optneur}. To prove Theorem~\ref{optneurtheo}(b), we observe that
each of the constituent functions being in $W_{r,2}^{\mbox{NN}}$,
(\ref{optneurest}) applied with $d=2$ implies that each of these
functions can be approximated from $\mathcal{S}_n$ up to accuracy
$n^{-r/2}$.  Our assumption that $f\in W_{H,r,2}^{\mbox{NN}}$ implies that  each of these constituent functions is Lipschitz
continuous. Hence, it is easy to deduce that, for example, if $P$, $P_1$,
$P_2$ are approximations to the constituent functions $h$, $h_1$,
$h_2$, respectively within an accuracy of $\epsilon$, then
$$
\|h(h_1,h_2)-P(P_1,P_2)\| \le c\epsilon,
$$
for some constant $c>0$ independent of the functions involved. This
leads to (\ref{deepnetapprox}).
$\square$

The constants involved
in $\O$ in (\ref{optneurest}) will depend upon the norms of the
derivatives of $f$ as well as $\sigma$. Thus, when the only a priori
assumption on the target function is about the number of derivatives,
then to \textbf{guarantee} an accuracy of $\epsilon$, we need a
shallow network with $\O(\epsilon^{-d/r})$ trainable parameters. If we assume a hierarchical structure on the target function as in Theorem~\ref{optneurtheo}, then the corresponding deep network yields a guaranteed accuracy of $\epsilon$ only with $\O(\epsilon^{-2/r})$ trainable parameters.

Is this the best? To investigate this question, let $M_n :W_{r,d}^{\mbox{NN}}\to \mathbb R^n$
be a continuous mapping (parameter selection), and $A_n :\mathbb R^n\to
C(I^d)$ be any mapping (recovery algorithm). Then an approximation to
$f$ is given by $A_n(M_n(f))$, where the continuity of $M_n$ means
that the selection of parameters is robust with respect to
perturbations in $f$. The
 nonlinear $n$--width of the compact set $W_{r,d}^{\mbox{NN}}$ is defined by
(cf.   \cite{devore1989optimal})
\begin{equation}
d_n(W_{r,d}^{\mbox{NN}})=\inf_{M_n, A_n}\sup_{f\in W_{r,d}^{\mbox{NN}}}\|f-A_n(M_n(f))\|,
\label{nwidthdef}
\end{equation}
and the \textit{curse for $W_{r,d}^{\mbox{NN}}$} by
\begin{equation}
\mathsf{curse}(W_{r,d}^{\mbox{NN}},\epsilon)=\min\{n\ge 1 : d_n(W_{r,d}^{\mbox{NN}})\le \epsilon\}.
\label{cursedef}
\end{equation}
We note that the curse depends only on the compact set $W_{r,d}^{\mbox{NN}}$, and represents the best
that can be achieved by \textbf{any} continuous parameter selection
and recovery processes.  It is shown by   \cite{devore1989optimal} that
$\mathsf{curse}(W_{r,d}^{\mbox{NN}},\epsilon)\ge c\epsilon^{-d/r}$ for some constant
$c>0$ depending only on $d$ and $r$. So, the estimate implied by
(\ref{optneurest}) is {\it the best possible} among \textbf{all} reasonable
methods of approximating arbitrary functions in $W_{r,d}^{\mbox{NN}}$, although by
itself, the estimate (\ref{optneurest}) is blind to the process by
which the approximation is accomplished; in particular, this process
is not required to be robust. Similar considerations apply to the estimate (\ref{deepnetapprox}), and we will explain the details in Section~\ref{gausssect} in a different context.

The lower bound on the $n$--width implies only that there is some
function in $W_{r,d}^{\mbox{NN}}$ for which the approximation cannot be better than that
suggested by (\ref{optneurest}). This begs the question whether this
function could be unreasonably pathological, and for most functions
arising in practice, clever ideas can lead to a substantially better
accuracy of approximation, its smoothness notwithstanding. At this
time, we are not able to address this question in the context of
neural networks as in Theorem~\ref{optneurtheo}, but we are able to do
so if each unit evaluates a Gaussian network. Accordingly, we now turn
to our new results in this direction.  The proofs will be published
separately.

\subsection{Deep and shallow Gaussian networks}\label{gausssect}
 
We wish to consider shallow networks where each channel evaluates a
Gaussian non--linearity; i.e., Gaussian networks of the form
\be\label{gaussnetworkdef} G(\x)=\sum_{k=1}^N
a_k\exp(-|\x-\x_k|^2),\qquad \x\in\mathbb R^d.  \ee It is natural to
consider the number of trainable parameters $(d+1)N$ as a measurement
of the complexity of $G$.  However, it is  known  (\cite{convtheo}) that an even more
important quantity that determines the approximation power of
Gaussian networks is the minimal separation among the centers.  For
any subset $\C$ of $\mathbb R^d$, the minimal separation of $\C$ is defined
by \be\label{minsepdef} \eta(\C)=\inf_{\x,\y\in\C, \x\not=\y}|\x-\y|.
\ee For $N, m>0$, the symbol $\mathcal{N}_{N,m}(\mathbb R^d)$ denotes the
set of all Gaussian networks of the form (\ref {gaussnetworkdef}),
with $\eta(\{\x_1,\cdots,\x_N\})\ge 1/m$.

Since it is our goal to make a comparison between shallow and deep
networks, we will consider also deep networks organized for simplicity
as a binary tree $\mathcal{T}$, where each unit computes a network in
$\mathcal{N}_{N,m}(\mathbb R^2)$. The set of all such networks will be
denoted by $\mathcal{T}\mathcal{N}_{N,m}(\mathbb R^2)$.  In this
context, it is not natural to constrain the output of the hidden units
to be in $[-1,1]$. Therefore, we need to consider approximation on the
entire Euclidean space in question.  Accordingly, let $\XX_d$ be the
space $C_0(\mathbb R^d)$ of continuous functions on $\mathbb R^d$
vanishing at infinity, equipped with the norm
$\|f\|_d=\max_{\x\in\mathbb R^d}|f(\x)|$. For the class $W$, we then
need to put conditions not just on the number of derivatives but also
on the rate at which these derivatives tend to $0$ at
infinity. Generalizing an idea from   \cite{freud1972direct,tenswt}, we
first define the space $W_{r,d}$ for integer $r\ge 1$ as the set of
all functions $f$ which have $r$ continuous derivatives in
$C_0(\mathbb R^d)$, satisfying
$$
\|f\|_{r,d}=\|f\|_d+\sum_{1\le|\k|_1\le r}\|\exp(-|\cdot|^2)D^\k(\exp(|\cdot|^2)f\|_d <\infty.
$$
 Since one of our goals is to show that our results on the upper
bounds for the accuracy of approximation are the best possible for
individual functions, the class $W_{r,d}$ needs to be refined
somewhat.  Toward that goal, we define next a regularization
expression, known in approximation theory parlance as a
$K$--functional,  by
$$
K_{r,d}(f,\delta)=\inf_{g\in W_{r,d}}\{\|f-g\|_d+\delta^r(\|g\|_d+\|g\|_{r,d})\}.
$$
We note that the infimum above is over \textbf{all} $g$ in the class
$W_{r,d}$ rather than just the class of all networks. The class
$\mathcal{W}_{\gamma,d}$ of functions which we are interested in is
then defined for $\gamma>0$ as the set of all $f\in C_0(\mathbb R^d)$ for
which
$$
\|f\|_{\gamma,d}=\|f\|+\sup_{\delta\in (0,1]}\frac{K_{r,d}(f,\delta)}{\delta^\gamma}<\infty,
$$
for some integer $r\ge \gamma$. It turns out that different choices of
$r$ yield equivalent norms, without changing the class itself.  The
following theorem gives a bound on approximation of $f\in C_0(\mathbb R^d)$
from $\mathcal{N}_{N,m}$. Here and in the sequel, we find it
convenient to adopt the following convention. By $A  \lesssim B$ we mean
$A\le cB$ where $c>0$ is a constant depending only on the fixed
parameters of the discussion, such as $\gamma$, $d$. By $A\sim B$, we
mean $A \lesssim B$ and $B \lesssim A$. 
The following theorems, as they apply to shallow networks, are technical improvements on those in \cite{convtheo}.

\begin{theorem}\label{unidegapptheo}
  There exists a constant $c>0$ depending on $d$ alone with the
  following property. Let $\{\C_m\}$ be a sequence of finite subsets
  with $\C_m\subset [-cm,cm]^d$, with \be\label{uniformity}
  1/m \lesssim\max_{\y\in K}\min_{\x\in \C}|\x-\y| \lesssim \eta(\C_m),
  \qquad m=1,2,\cdots.  \ee If $\gamma>0$ and $f\in
  \mathcal{W}_{\gamma,d}$, then for integer $m\ge 1$, there exists
  $G\in \mathcal{N}_{|\C_m|,m}$ with centers at points in $\C_m$ such
  that \be\label{unidirect} \|f-G\|_d \lesssim
  \frac{1}{m^\gamma}\|f\|_{\gamma,d}.  \ee Moreover, the coefficients
  of $G$ can be chosen as linear combinations of the data $\{f(\x)
  :\x\in\C_m\}$.
\end{theorem}

We note that the set of centers $\C_m$ can be chosen arbitrarily
subject to the conditions stated in the theorem; \textbf{there is no
  training necessary to determine these parameters}. Therefore, there
are only $\O(m^2)$ coefficients to be found by training. This means
that if we assume a priori that $f\in \mathcal{W}_{\gamma,d}$, then
the number of trainable parameters to theoretically guarantee an
accuracy of $\epsilon>0$ is $\O(\epsilon^{-2d/\gamma})$. We will
comment on the optimality of this estimate later.

Next, we discuss approximation by deep networks in
$\mathcal{T}\mathcal{N}_{N,m}(\mathbb R^2)$. We will show that the accuracy
of the approximation increases dramatically if the target function $f$
is known to have the compositional hierarchical structure prescribed
by $\mathcal{T}$. It is not clear that this structure is
unique. Therefore, for mathematical analysis, it is convenient to
think of such a function $f$ as in fact a family of functions
$\{f_v\}_{v\in V}$, where $V$ is the set of non--leaf nodes in
$\mathcal{T}$, and $f_v$ is the constituent function evaluated at the
vertex $v$. The set of all such functions will be denoted by
$\mathcal{T}C_0(\mathbb R^2)$.  Likewise, a network
$G\in\mathcal{T}\mathcal{N}_{N,m}(\mathbb R^2)$ is thought of as the family
of networks $\{G_v\}_{v\in V}$, where each $G_v\in
\mathcal{N}_{N,m}(\mathbb R^2)$.  Accordingly, the norm in which the
approximation error (respectively, the smoothness) is measured is
defined by \be\label{treenormdef} \|f\|_{\mathcal{T}}=\sum_{v\in
  V}\|f_v\|_2, \qquad
\|f\|_{\mathcal{T}\mathcal{W}_{\gamma,2}}=\sum_{v\in
  V}\|f_v\|_{\mathcal{W}_{\gamma,2}}.  \ee The analogue of
Theorem~\ref{unidegapptheo} is the following.
\begin{theorem}
\label{treedegapptheo}
For each $v\in V$, let $\{\C_{m,v}\}$ be a sequence of finite subsets
as described in Theorem~\ref{unidegapptheo}. Let $\gamma>0$ and
$f\in\mathcal{T}\mathcal{W}_{\gamma,2}$. Then for integer $m\ge 1$,
there exists $G\in \mathcal{T}\mathcal{N}_{\max|\C_{m,v}|,m}(\mathbb R^2)$
with centers of the constituent network $G_v$ at vertex $v$ at points
in $\C_{m,v}$ such that \be\label{gfuncdirect}
\|f-G\|_{\mathcal{T}} \lesssim
\frac{1}{m^\gamma}\|f\|_{\mathcal{T},\gamma,2}.  \ee Moreover, the
coefficients of each constituent $G_v$ can be chosen as linear
combinations of the data $\{f(\x) :\x\in\C_{m,v}\}$.
\end{theorem}

Clearly, Theorem~\ref{treedegapptheo} is applicable only for those
target functions which have the hierarchical structure prescribed by
the binary tree. It is not difficult to generalize the theorem to the
case when the structure confirms rather to a more general directed
acyclic graph, but for simplicity, we will continue to assume the
binary tree structure in this section. Therefore,
Theorem~\ref{treedegapptheo} can be compared with
Theorem~\ref{unidegapptheo} only in the following sense. A target
function $f$ satisfying the tree structure can also be thought of as a
\textit{shallow} function of $L$ arguments, where $L$ is the number of
leaves of the binary tree (the input variables). Then the set $V$ contains $\le L$ elements as
well. If $f$ satisfies the smoothness conditions in both the theorems,
and an accuracy of $\epsilon>0$ is required, then a shallow network
requires $\O(\epsilon^{-2L/\gamma})$ trainable parameters,
while the deep network requires only $\O(L\epsilon^{-4/\gamma})$
trainable parameters.
 
How good are these results for individual functions? If we know that
some oracle can give us Gaussian networks that achieve a given
accuracy with a given complexity, does it necessarily imply that the
target function is smooth as indicated by the above theorems? It is in
this context that we need to measure the complexity in terms of
minimal separation among the centers; such a result will then hold
even if we allow the oracle to chose a very large number of
channels. The following is a converse to Theorems~\ref{unidegapptheo}
and \ref{treedegapptheo}, demonstrating that the accuracy asserted by
these theorems is possible if and only if the target function is in
the smoothness class required in these theorems.

\begin{theorem}
  \label{convtheo} {\rm (a)} Let $\{\C_m\}$ be a sequence of finite
  subsets of $\mathbb R^d$, such that for each integer $m\ge 1$,
  $\C_m\subseteq \C_{m+1}$, $|\C_m|\le c\exp(c_1m^2)$, and
  $\eta(\C_m)\ge 1/m$. Further, let $f\in C_0(\mathbb R^d)$, and for each
  $m\ge 1$, let $G_m$ be a Gaussian network with centers among points
  in $\C_m$, such that \be\label{uniconv_degapprox} \sup_{m\ge
    1}m^\gamma\|f-G_m\|_d <\infty.  \ee
  Then $f\in \mathcal{W}_{\gamma,d}$. \\
  {\rm (b)} For each $v\in V$, let $\{\C_{m,v}\}$ be a sequence of
  finite subsets of $\mathbb R^{d(v)}$, satisfying the conditions as
  described in part (a) above. Let $f\in\mathcal{T}C_0(\mathbb R^2)$,
  $\gamma>0$, and $\{G_m\in \mathcal{T}\mathcal{N}_{n,m}$\} be a
  sequence where, for each $v\in V$, the centers of the constitutent
  networks $G_{m,v}$ are among points in $\C_{m,v}$, and such that
  \be\label{gfuncconv_degapprox} \sup_{m\ge
    1}m^\gamma\|f-G_m\|_{\mathcal{T}} <\infty.  \ee Then $f\in
  \mathcal{T}\mathcal{W}_{\gamma,2}$.

\end{theorem}

\subsection{VC bounds}

A direct connection between regression and binary classification is
provided by the following observation (due to
\cite{DBLP:journals/corr/abs-1304-7045}): Theorems 11.13 and 14.1 from
\cite{AntBartlett2002} show that the fat-shattering dimension is
upper-bounded by the VC-dimension of a slightly larger class of
networks, which has a similar VC-dimension to the original class,
hence the fat-shattering dimension can also be bounded. The following
theorem can be deduced from the results in Section~\ref{deepnetapprox}
and a well known result (\cite{AntBartlett2002}, \cite{Mhaskaretal2016}):

\begin{theorem}\label{vcdimtheo}
The VC-dimension of the shallow network with
    $N$ units is bounded
    by $(d+2)N^2$; the VC-dimension of the  binary tree network with
    $n (d-1)$ units is bounded by $4 n^2 (d-1)^2$.
\end{theorem}

\section{A general framework for hierarchical, compositional
  computations}

There are many phenomena in nature that have descriptions along a
range of rather different scales. An extreme case consists of fractals
which are infinitely self-similar, iterated mathematical
constructs. As a reminder, a self-similar object is similar to a part
of itself (i.e. the whole is similar to one or more of the
parts). Many objects in the real world are statistically self-similar,
showing the same statistical properties at many scales: clouds, river
networks, snow flakes, crystals and neurons branching.  A relevant
point is that the shift-invariant scalability of image statistics
follows from the fact that objects contain smaller clusters of similar
surfaces in a selfsimilar fractal way. \cite{Ruderman1997}
analysis  shows that image statistics reflects
the property of compositionality of objects and parts: parts are
themselves objects, that is selfsimilar clusters of similar surfaces
in the physical world. The closely related property of {\it
  compositionality} was a main motivation for hierarchical models of
visual cortex such as HMAX which can be regarded as a pyramid of AND
and OR layers (\cite{Riesenhuber1999}), that is a sequence of
conjunctions and disjunctions.

The architecture of algorithms that are applied to data characterized
by many scales -- such as images -- should exploit this property in
their architecture. A way to formulate the requirements on the
algorithms is to assume that {\it scalable computations} are a
subclass of nonlinear discrete operators, mapping vectors from $\R^n$
into $\R^d$ (for simplicity we put in the following $d=1$). Informally
we call an algorithm $K_n: \R^n \mapsto \R$ {\it scalable} if it maintains
the same ``form'' when the input vectors increase in dimensionality;
that is, the same kind of computation takes place when the size of the
input vector changes. This motivates the following construction and
definitions. Consider a ``layer'' operator $H_{2m}: \R^{2m} \mapsto
\R^{2m-2}$ for $m \ge 1$ with a special structure that we call ``shift
invariance''.

\begin{definition}\label{scalabledef}
  For integer $m \ge 2$, an operator $H_{2m}$ is shift-invariant if
  $H_{2m} = H_m' \oplus H_m''$ where $\R^{2m}=\R^m \oplus \R^m$,
  $H'=H''$ and $H':\R^m \mapsto \R^{m-1}$.  An operator $K_{2M}
  :\R^{2M}\to \R$ is called scalable and shift invariant if
  $K_{2M}=H_2\circ \cdots H_{2M}$ where each $H_{2k}$, $1\le k\le M$,
  is shift invariant.
\end{definition}

\begin{figure}\centering
\includegraphics[width=0.5\textwidth]{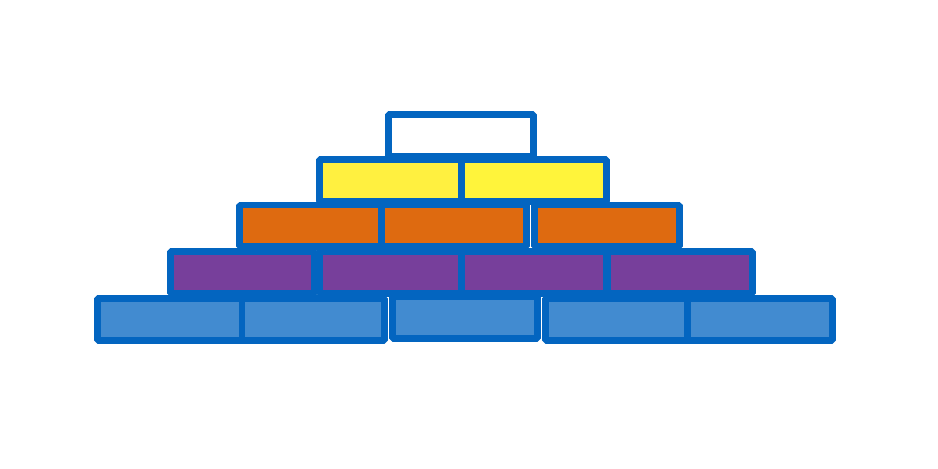}
\caption{A scalable operator.  Each
layer consists of identical blocks; each block is an operator $H_2: \R^2
\mapsto R$}
\label{ScalableOperator} 
\end{figure}

 We observe that \textit{ scalable shift-invariant operators $K: \R^{2m}
  \mapsto \R$ have the structure $K = H_2 \circ H_4 \circ H_6 \cdots
  \circ H_{2m}$, with $H_4=H'_2 \oplus H'_2$, $H_6=H''_2 \oplus H''_2 \oplus
  H_2''$, etc.}.
 Thus the structure of a {\it shift-invariant, scalable
  operator} consists of several layers; each layer consists of
identical blocks; each block is an operator $H: \R^2 \mapsto R$ (See Figure \ref{ScalableOperator}). We
note also that an alternative weaker constraint on $H_{2m}$ in
Definition~\ref{scalabledef}, instead of shift invariance, is mirror
symmetry, that is $H''= R \circ H'$, where $R$ is a
reflection. Obviously, shift-invariant scalable operator are
equivalent to shift-invariant compositional functions.

The final step in the argument uses the results of previous sections
to claim that a nonlinear node with two inputs and enough units (that
is, channels) can approximate arbitrarily well each of the $H_2$
blocks. This leads to conclude that deep convolutional neural networks are natural
approximators of {\it scalable, shift-invariant operators}.

\section{Discussion}

Implicit in the results in  Section~\ref{degapproxsect} is the fact that a
hierarchical network can approximate a high degree polynomial $P$
in the input variables $x_1,\cdots,x_d$, that can be written as a
hierarchical composition of lower degree polynomials.  For example,
let
\begin{eqnarray*}
Q(x,y)&\!\!\!=\!\!\!&(Ax^2y^2+Bx^2y+Cxy^2+Dx^2+2Exy+  Fy^2+2Gx+2Hy+I)^{2^{10}}.
\end{eqnarray*}

\begin{figure*}[ht]
\centering
\includegraphics[width=0.8\textwidth]{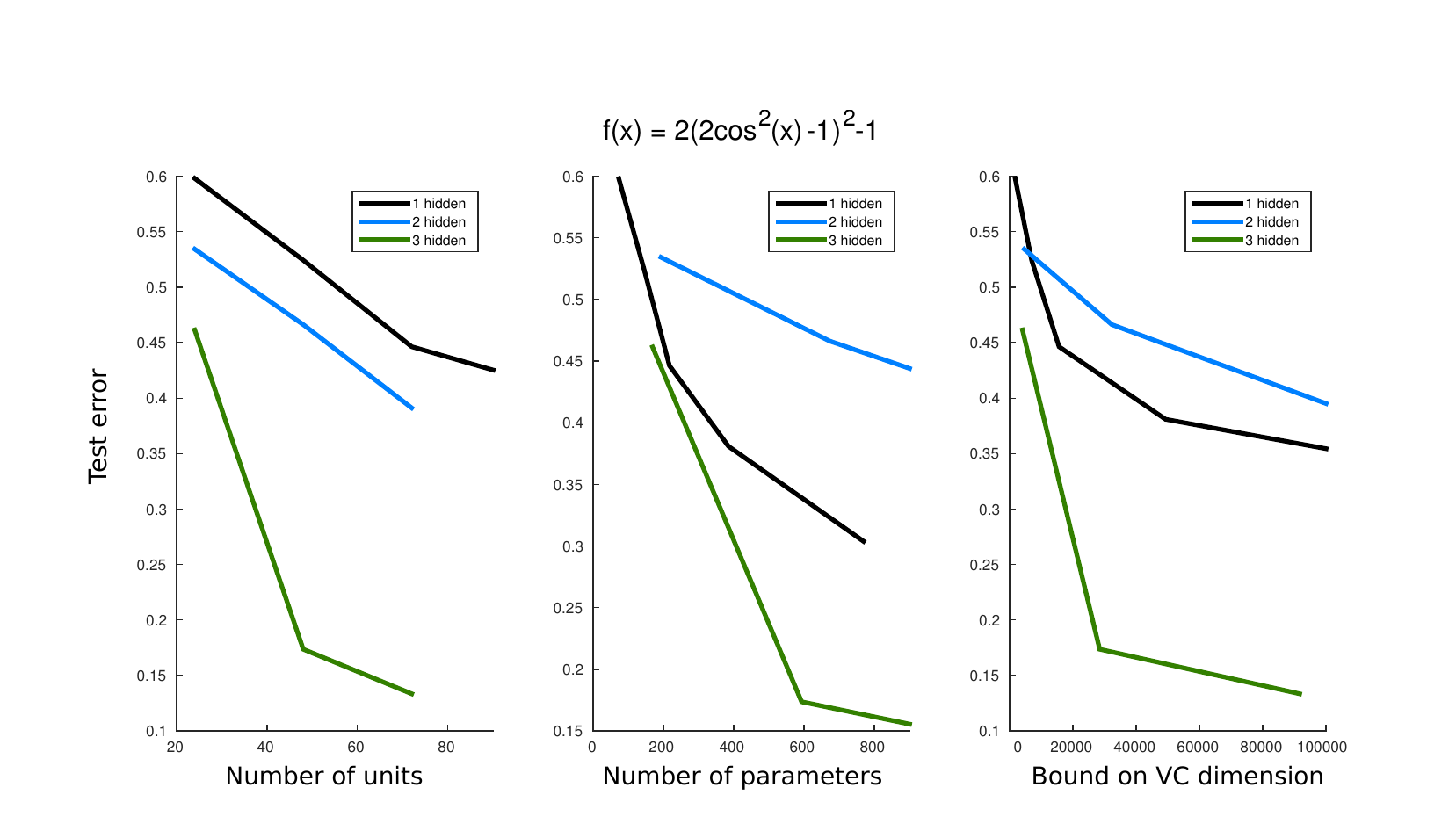}
\caption{A sparse trigonometric function $f(x)=2(2\cos^2(x)-1)^2-1$ with
  one input variable is learned in a regression
  set-up using standard deep networks with 1, 2 or 3 hidden layers. In
  the 1 hidden layer setting, 24, 48, 72, 128 and 256 hidden units
  were tried. With 2 hidden layers, 12, 24 and 36 units per layer were
  tried. With 3 hidden layers, 8, 16 and 24 units per layer were
  tried. Each of the above settings was repeated 5 times, reporting
  the lowest test error. Mean squared error (MSE) was used as the
  objective function; the y axes in the above figures are the square
  root of the testing MSE. For the experiments with 2 and 3 hidden
  layers, batch normalization   \cite{ioffe2015batch} was used between
  every two hidden layers. 60k training and 60k testing samples were
  drawn from a uniform distribution over $[-2\pi, 2\pi]$. The training
  process consisted of 2000 passes through the entire training data
  with mini batches of size 3000. Stochastic gradient descent with
  momentum 0.9 and learning rate 0.0001 was used. Implementations were
  based on MatConvNet   \cite{vedaldi2015matconvnet}. Same data points
  are plotted in 2 sub-figures with x axes being number of units and
  parameters, respectively. Note that with the input being 1-D, the
  number of parameters of a shallow network scales slowly with respect
  to the number of units, giving a shallow network some advantages in
  the right sub-figure. Although not shown here, the training errors
  are very similar to those of testing. The advantage of deep networks
  is expected to increase with increasing dimensionality of the
  function. Even in this simple case the solution found by SGD are almost
  certain to be suboptimal. Thus the figure cannot be taken as fully
  reflecting the theoretical results of this paper.}
\label{cos_4th}
\end{figure*}

Since $Q$ is nominally a polynomial of coordinatewise degree $2^{11}$,
  \cite[Lemma~3.2]{optneur} shows  that a shallow network with
$2^{11}+1$ units is able to approximate $Q$ arbitrarily well on
$I^d$. However, because of the hierarchical structure of $Q$,
  \cite[Lemma~3.2]{optneur} shows also that a hierarchical network
with $9$ units can approximate the quadratic expression, and $10$
further layers, each with $3$ units can approximate the successive
powers. Thus, a hierarchical network with $11$ layers and $39$ units
can approximate $Q$ arbitrarily well. We note that even if $Q$ is
nominally of degree $2^{11}$, each of the monomial coefficients in $Q$
is a function of only $9$ variables, $A,\cdots, I$. A similar, simpler
example was tested using standard DLNN software and is shown in Figure
\ref{cos_4th}.

These arguments suggest that the proof of Theorem 1 can be used to
show (see   \cite{Mhaskaretal2016}) that functions approximated well by
sparse polynomials can be learned efficiently by deep networks with a
tree or graph structure that matches the polynomial.  We recall that in a
similar way several properties of certain Boolean functions can be
``read out'' from the terms of their Fourier expansion corresponding
to ``large'' coefficients, that is from a polynomial that approximates
well the function (see \cite{poggio2015December}). In this sense our Theorem 1 should cover recently
described functions that cannot be represented efficiently by shallow
networks (see   \cite{Telgarsky2015}).

Classical results (\cite{Hastad1987}) about the depth-breadth tradeoff
in circuits design show that deep circuits are more efficient in
representing certain Boolean functions than shallow circuits. These
results  have been often quoted in support of the
claim that deep neural networks can represent functions that shallow
networks cannot. For instance \cite{BengioLecun2007} write {\it ``We
  claim that most functions that can be represented compactly by deep
  architectures cannot be represented by a compact shallow
  architecture''}. The results of this paper (see Supplementary Material and
\cite{Mhaskaretal2016}) should settle the issue, justifying the
original conjecture and providing an approach connecting results on
Boolean functions with current real valued neural networks.

\bibliographystyle{apalike}

\small
\bibliography{Boolean}
\normalsize

\appendix

%\section{Scalable Algorithms}
% Figure \ref{ScalableOperator} refers to the main text on scalable operators.

\section{Boolean functions}

Our results sketched in the previous section are interesting not only
in themselves but also because they suggest several connections to
similar properties of Boolean functions. In fact our results seem to
generalize properties already known for Boolean functions which are of
course a special case of functions of real variables. We first recall some
definitions followed by a few  observations.

One of the most important and versatile tools for theoretical computer
scientists for the study of functions of $n$ Boolean variables, their
related circuit design and several associated learning problems, is
the Fourier transform over the Abelian group $\mathcal{Z}^n_2$ . This
is known as Fourier analysis over the Boolean cube $\{-1,1\}^n$. The
Fourier expansion of a Boolean function $f: \{-1,1 \}^n \to \{-1,1\}$
or even a real-valued Boolean function $f: \{-1,1 \}^n \to [-1,1]$ is
its representation as a real polynomial, which is multilinear because
of the Boolean nature of its variables. Thus for Boolean functions
their Fourier representation is identical to their polynomial
representation. In the following we will use the two terms
interchangeably. Unlike functions of real variables, the full finite
Fourier expansion is exact instead of an approximation and there is no
need to distingush between trigonometric and real polynomials. Most of
the properties of standard harmonic analysis are otherwise preserved,
including Parseval theorem. The terms in the expansion correspond to
the various monomials; the low order ones are parity functions over
small subsets of the variables and correspond to low degrees and low
frequencies in the case of polynomial and Fourier approximations,
respectively, for functions of real variables.

The section in the main text referring to sparse functions suggests the following approach to
characterize which functions are best learned by which type of network
-- for instance shallow or deep. The structure of the network is
reflected in polynomials that are best approximated by it -- for
instance generic polynomials or sparse polynomials (in the
coefficients) in $d$ variables of order $k$. The tree structure of the
nodes of a deep network reflects the structure of a specific sparse
polynomial. Generic polynomial of degree $k$ in $d$ variables are
difficult to learn because the number of terms, trainable parameters
and associated VC-dimension are all exponential in $d$. On the other
hand, functions approximated well by sparse polynomials can be learned
efficiently by deep networks with a tree structure that matches the
polynomial.  We recall that in a similar way several properties of
certain Boolean functions can be ``read out'' from the terms of their
Fourier expansion corresponding to ``large'' coefficients, that is
from a polynomial that approximates well the function.

Classical results \cite{Hastad1987} about the depth-breadth tradeoff
in circuits design show that deep circuits are more efficient in
representing certain Boolean functions than shallow circuits. Hastad
proved that highly-variable functions (in the sense of having high
frequencies in their Fourier spectrum) in particular the parity
function cannot even be decently approximated by small constant depth
circuits (see also \cite{LinialMansour1993}). These results on Boolean
functions have been often quoted in support of the claim that deep
neural networks can represent functions that shallow networks
cannot. For instance Bengio and LeCun \cite{BengioLecun2007} write
{\it ``We claim that most functions that can be represented compactly by
  deep architectures cannot be represented by a compact shallow
  architecture''.''}. It seems
that the results summarized in this paper provide a general approach
connecting results on Boolean functions with current real valued
neural networks. Of course, we do not imply that the capacity of deep
networks is exponentially larger than the capacity of shallow
networks. As pointed out by Shalev-Shwartz, this is clearly not true,
since the VC dimension of a network depends on the number of nodes and
parameters and not on the depth. We remark that a nice theorem was
recently published \cite{Telgarsky2015}, showing that a certain family
of classification problems with real-valued inputs cannot be
approximated well by shallow networks with fewer than exponentially
many nodes whereas a deep network achieves zero error. This is a
special case of our results and  corresponds to high-frequency, sparse
trigonometric polynomials.

Finally, we want to speculate about a series of observations on
Boolean functions that may show an interesting use of our approach
using the approximating polynomials and networks for studying the
learning of general functions. It is known that within Boolean
functions the $AC^0$ class of polynomial size constant depth circuits
is characterized by Fourier transforms where most of the power
spectrum is in the low order coefficients. Such functions can be
approximated well by a polynomial of low degree and can be learned
well by considering only such coefficients. In general, two algorithms
\cite{Mansour1994} seems to allow learning of certain Boolean function
classes:
\begin{enumerate}
\item the low order algorithm that approximates functions by
  considering their low order Fourier coefficients and
\item the sparse algorithm which learns a function by
  approximating its significant coefficients.
\end{enumerate}
Decision lists and decision trees can be learned by algorithm
1. Functions with small $L_1$ norm can be approximated well by
algorithm 2. Boolean circuits expressing DNFs can be approximated by 1
but even better by 2. In fact, in many cases most of the coefficients
of the low terms may still be negligeable and furthermore it may the
case that a function can be approximated by a small set of
coefficients but these coefficients do not correspond to low-order
terms. All these cases are consistent with the description we have in
section on sparse functions. For general functions they may suggest
the following. Many functions can be learned efficiently in terms of
their low order coefficients and thus by shallow networks. This
corresponds to using Tikhonov regularization that effectively cuts out
high frequencies. Other functions must be learned in terms of their
sparse coefficients by a deep network with an appropriate
architecture. This is more similar to $L_1$ regularization. The
sparsity approach which corresponds to deep networks includes the
shallow Tikhonov approach and thus is more general and preferrable at
least as long as computational and sample complexity issues are not
taken into account.

\end{document}